\pdfoutput=1

\documentclass[11pt]{article}

\usepackage[final]{acl}

\usepackage{times}
\usepackage{latexsym}

\usepackage[T1]{fontenc}

\usepackage[utf8]{inputenc}

\usepackage{microtype}

\usepackage{inconsolata}

\usepackage{graphicx}

\usepackage{tikz}
\usepackage{tikz}
\usepackage[edges]{forest}
\definecolor{hidden-draw}{RGB}{205, 44, 36}
\definecolor{hidden-blue}{RGB}{194,232,247}
\definecolor{hidden-orange}{RGB}{243,202,120}
\definecolor{hidden-yellow}{RGB}{242,244,193}
\definecolor{tree-level-1}{RGB}{245,20,85}
\definecolor{tree-level-2}{RGB}{246,86,118}
\definecolor{tree-level-3}{RGB}{248,177,193}
\definecolor{tree-leaf}{RGB}{176,230,198}

\usepackage{scalerel}

\usepackage{array}
\usepackage{booktabs}
\usepackage{colortbl}
\usepackage{xcolor}
\usepackage{pifont}

\title{A Survey of Large Language Models in Psychotherapy: \\ Current Landscape and Future Directions}

\author{
    Hongbin Na\hspace{.05em}\scalerel*{\includegraphics{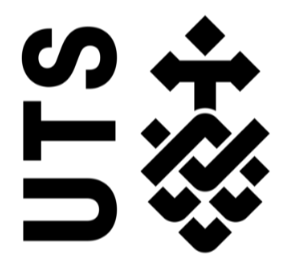}}{\textbf{ O}}\hspace{.05em}\thanks{Equal contribution.}, 
    Yining Hua\hspace{.05em}\scalerel*{\includegraphics{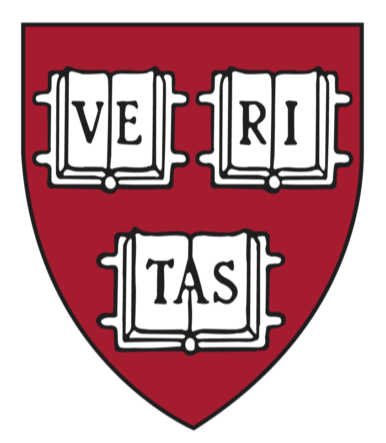}}{\textbf{O}}\hspace{.05em}\footnotemark[1], 
    Zimu Wang\hspace{.05em}\scalerel*{\includegraphics{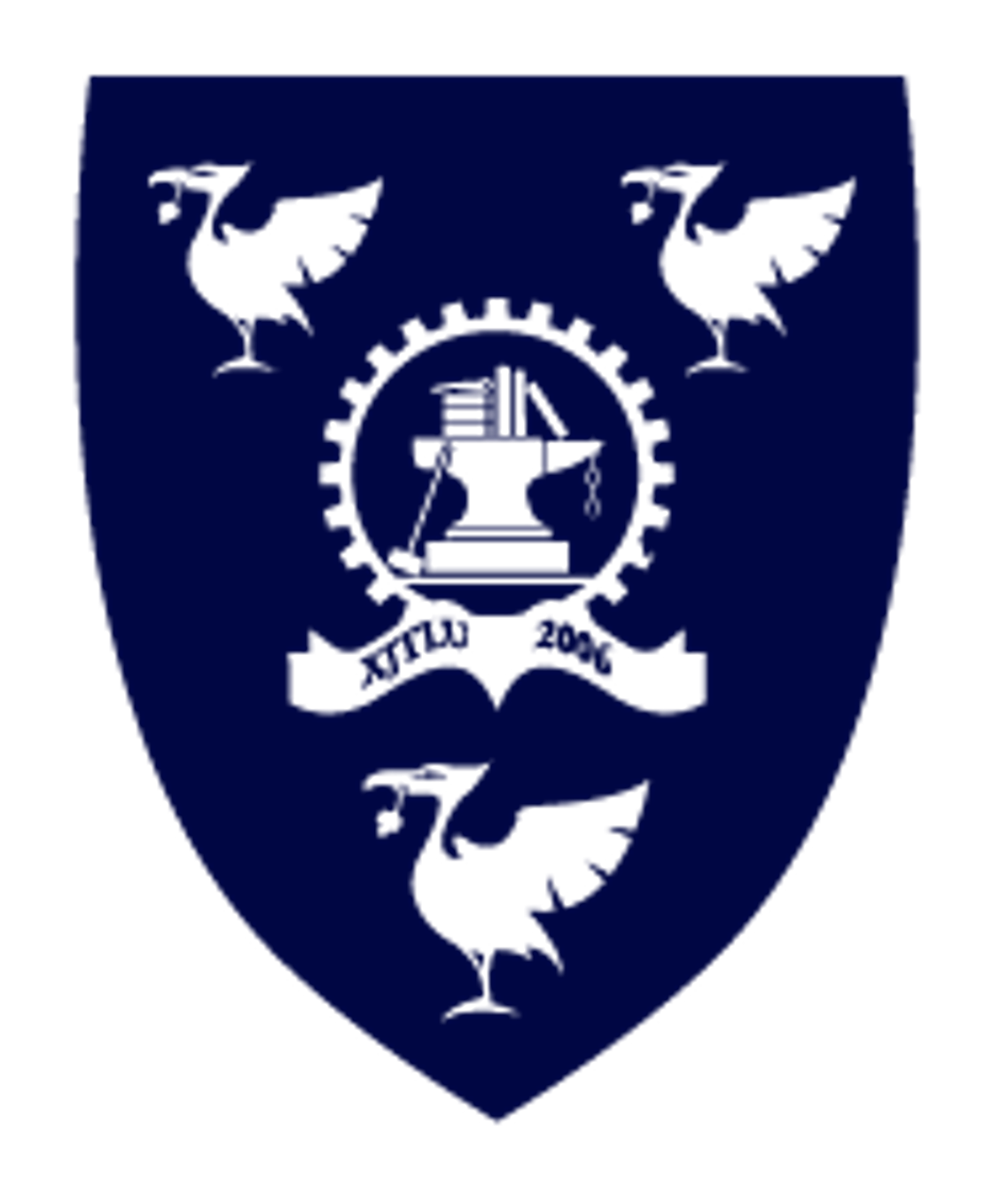}}{\textbf{O}}\hspace{.05em}\footnotemark[1],\\ 
    \textbf{Tao Shen\hspace{.05em}\scalerel*{\includegraphics{UTS_Logo.png}}{\textbf{O}}\hspace{.05em}, 
    Beibei Yu\hspace{.05em}\scalerel*{\includegraphics{UTS_Logo.png}}{\textbf{O}}\hspace{.05em}, 
    Lilin Wang\hspace{.05em}\scalerel*{\includegraphics{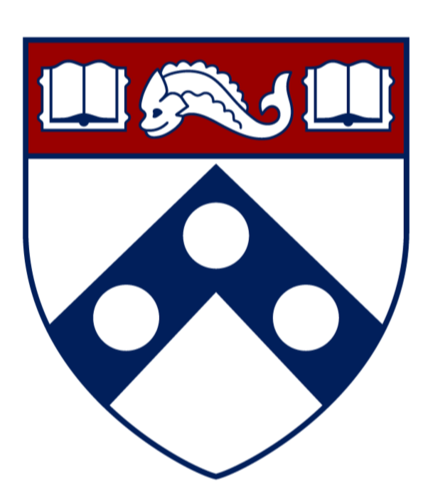}}{\textbf{O}}\hspace{.05em}, 
    Wei Wang\hspace{.05em}\scalerel*{\includegraphics{XJTLU_Logo.png}}{\textbf{O}}\hspace{.05em}, 
    John Torous\hspace{.05em}\scalerel*{\includegraphics{Harvard_Logo.png}}{\textbf{O}}\hspace{.05em}, 
    Ling Chen\hspace{.05em}\scalerel*{\includegraphics{UTS_Logo.png}}{\textbf{O}}\hspace{.05em}} \\ 
    \hspace{.05em}\scalerel*{\includegraphics{UTS_Logo.png}}{\textbf{O}}\hspace{.05em}Australian Artificial Intelligence Institute, University of Technology Sydney \\ 
    \hspace{.05em}\scalerel*{\includegraphics{Harvard_Logo.png}}{\textbf{O}}\hspace{.05em}Harvard University \ \ 
    \hspace{.05em}\scalerel*{\includegraphics{XJTLU_Logo.png}}{\textbf{O}}\hspace{.05em}Xi'an Jiaotong-Liverpool University \ \ 
    \hspace{.05em}\scalerel*{\includegraphics{Penn_Logo.png}}{\textbf{O}}\hspace{.05em}University of Pennsylvania \\ 
    \texttt{hongbin.na@student.uts.edu.au,  yininghua@g.harvard.edu} \\
    \texttt{zimu.wang19@student.xjtlu.edu.cn}
}

\begin{document}
\maketitle
\begin{abstract}
Mental health is increasingly critical in contemporary healthcare, with psychotherapy demanding dynamic, context-sensitive interactions that traditional NLP methods struggle to capture. Large Language Models (LLMs) offer significant potential for addressing this gap due to their ability to handle extensive context and multi-turn reasoning. This review introduces a conceptual taxonomy dividing psychotherapy into interconnected stages--assessment, diagnosis, and treatment--to systematically examine LLM advancements and challenges. Our comprehensive analysis reveals imbalances in current research, such as a focus on common disorders, linguistic biases, fragmented methods, and limited theoretical integration. We identify critical challenges including capturing dynamic symptom fluctuations, overcoming linguistic and cultural biases, and ensuring diagnostic reliability. Highlighting future directions, we advocate for continuous multi-stage modeling, real-time adaptive systems grounded in psychological theory, and diversified research covering broader mental disorders and therapeutic approaches, aiming toward more holistic and clinically integrated psychotherapy LLMs systems.
\end{abstract}

\section{Introduction}

Mental health plays an increasingly critical role in current healthcare and social well-being.
The high prevalence of common psychological disorders, such as depression and anxiety, has led to a growing demand for accessible and effective psychotherapy. The core of psychotherapy resides in \textit{dynamic, contextual} interpersonal interactions—therapists should continuously assess and adjust their intervention strategies~\cite{wampold2015great} based on patients' emotional fluctuations, verbal expressions, and social backgrounds, fostering a strong therapeutic alliance~\cite{stubbe2018therapeutic} to achieve symptom resilience. This deep and flexible process contrasts sharply with traditional NLP, which is typically limited to static or single-task settings.

\begin{figure}[t!]
\centering
\includegraphics[width=0.5\textwidth]{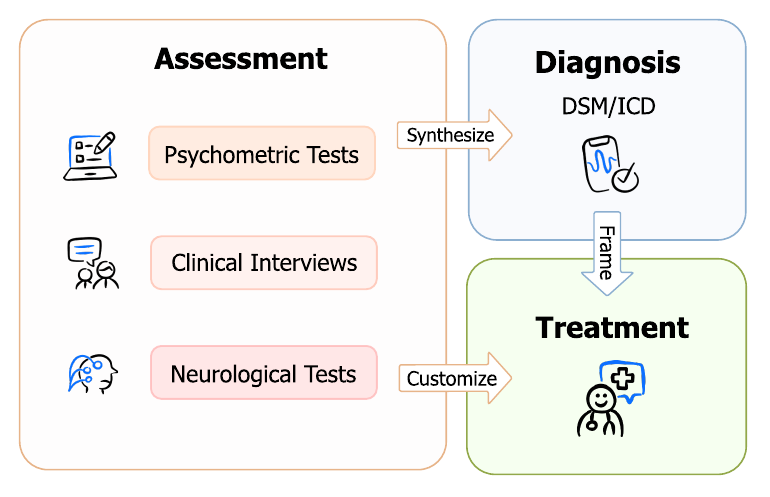}
\caption{The dynamic and interrelated network among assessment, diagnosis, and treatment in psychotherapy.}
\label{fig:Taxonomy}
\end{figure}

Large language models (LLMs) offer a new perspective to addressing this challenge. By leveraging their capability to model extensive context and perform multi-turn reasoning \cite{wang-etal-2024-document-level,NEURIPS2024_e560a0b2}, LLMs can capture rich semantics and emotional signals in dialogues \cite{ma-etal-2025-detecting}, enabling end-to-end language understanding and generation \cite{wang-etal-2024-knowledge,qian-etal-2024-domain}.
In assessment, LLMs can extract potential symptom cues from vague and fragmented expressions~\cite{24,108}. During diagnosis, they integrate subjective and objective patient information across multiple utterances~\cite{9,52}. In therapeutic interventions, they adapt conversational strategies based on patients' real-time feedback, enabling more flexible and human-like interactions compared to traditional scripted systems~\cite{40,78}. As a result, LLMs have the potential to surpass the conventional ``discrete label recognition'' paradigm, evolving toward a model of continuous, progressive clinical reasoning, enabling seamless connections across \textit{assessment}, \textit{diagnosis}, and \textit{treatment}, aligning more closely with therapists' cognitive process and interaction flow.

\definecolor{AssessmentColor}{HTML}{E7B489}  
\definecolor{DiagnosisColor}{HTML}{8CAFD0}   
\definecolor{TreatmentColor}{HTML}{BECD8F}

\tikzstyle{my-box}=[
    rectangle,
    rounded corners,
    text opacity=1,
    minimum height=1.5em,
    minimum width=5em,
    inner sep=2pt,
    align=center,
    fill opacity=.5,
]
\tikzstyle{leaf}=[my-box, minimum height=1.5em,
    text=black, align=left,font=\scriptsize,
    inner xsep=2pt,
    inner ysep=4pt,
]

\forestset{
  color subtree/.style={
    fill=#1!10,    
    draw=#1,       
    edge+={draw=darkgray, line width=1pt},
    for children={
      color subtree=#1   
    }
  }
}

\begin{figure*}[t]
    \centering
    \resizebox{\textwidth}{!}{
        \begin{forest}
            forked edges,
            for tree={
                grow=east,
                reversed=true,
                anchor=base west,
                parent anchor=east,
                child anchor=west,
                base=left,
                font=\small,
                rectangle,
                draw=hidden-draw,
                rounded corners,
                align=left,
                minimum width=4em,
                edge+={darkgray, line width=1pt},
                s sep=3pt,
                inner xsep=2pt,
                inner ysep=3pt,
                ver/.style={rotate=90, child anchor=north, parent anchor=south, anchor=center},
            },
            where level=1{text width=4.0em,font=\scriptsize,}{},
            where level=2{text width=5.6em,font=\scriptsize,}{},
            where level=3{text width=6.2em,font=\scriptsize,}{},
            where level=4{text width=6.0 em,font=\scriptsize,}{},
            [ LLMs in \\Psychotherapy
                [
                    Assessment,
                    color subtree=AssessmentColor,
                    [
                        Symptoms
                        [
                            Detection
                            [
                                \citet{22,24,27,66,95}; \\ \citet{85,91,94,96}; \\ \citet{108,120,18}
                                , leaf, text width=26.6em
                            ]
                        ]
                        [
                            Severity
                            [
                                \citet{11,23,58,100}; \\ \citet{59}
                                , leaf, text width=26.6em
                            ]
                        ]
                    ]
                    [
                        Cognition
                        [
                           \citet{5,10,12,14,16,38}\\ \citet{45}
                            , leaf, text width=34.5em
                        ]
                    ]
                    [
                        Behavior
                        [
                            \citet{36,60,65,67}
                            , leaf, text width=34.5em
                        ]
                    ]
                    [
                        Advanced
                        [
                            \citet{48,53,99}
                            , leaf, text width=34.5em
                        ]
                    ]
                ]
                [
                    Diagnosis,
                    color subtree=DiagnosisColor
                    [
                        Static Diagnosis
                        [
                            \citet{11,107,102,124,110}
                            , leaf, text width=34.5em
                        ]
                    ]
                    [
                        Dynamic Diagnosis
                        [
                            \citet{9,33,52,106}
                            , leaf, text width=34.5em
                        ]
                    ]
                ]
                [
                    Treatment,
                    color subtree=TreatmentColor
                    [
                        LLM as a Virtual \\ Therapist
                        [
                            \citet{20,29,40,54,55}; \\ \citet{77,78,109}
                            , leaf, text width=34.5em
                        ]
                    ]
                    [
                        LLM as an Assistive \\Tool
                        [
                            \citet{2,3,5,35,43}; \\ \citet{64,70,81}
                            , leaf, text width=34.5em
                        ]
                    ]
                    [
                        LLM as Simulated \\Patients for Clinician \\Education
                        [
                            \citet{49,51,56,80}
                            , leaf, text width=34.5em
                        ]
                    ]
                    [
                        LLM for Evaluation \\and Quality Analysis
                        [
                            \citet{4,19,32,37,39,62}; \\ \citet{65,67,101,103,122}
                            , leaf, text width=34.5em
                        ]
                    ]
                ]
            ]
        \end{forest}
    }
    \caption{Taxonomy of Research on Large Language Models in Psychotherapy.}
    \label{fig:overall}
\end{figure*}
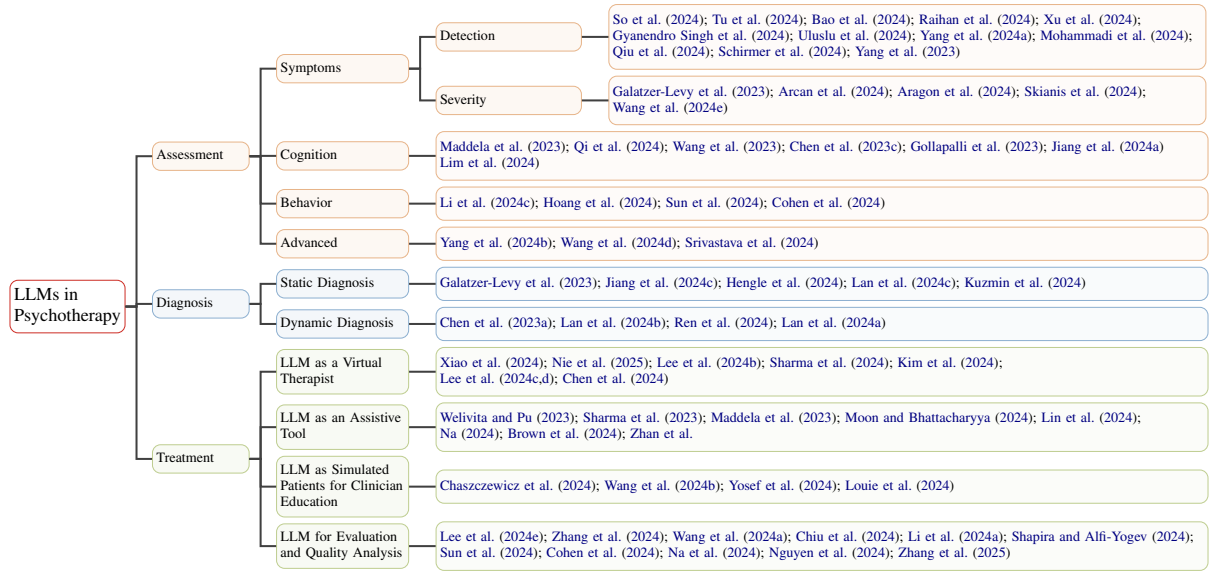

However, existing research on applying LLMs in this field remains somewhat \textit{fragmented}. Many studies have utilized LLMs for isolated tasks, such as depression detection~\cite{18,27} or diagnosis~\cite{107}, regarding them as superior feature extractors. Another research line has focused on developing mental health counseling chatbots \cite{chen-etal-2023-soulchat,19}; however, these systems remain limited to partial assistance due to insufficient integration with clinical workflows. In other words, although LLMs hold the potential to span the entire continuum from assessment to intervention, they remain limited by the fragmented paradigms of traditional NLP, preventing them from fully leveraging their dynamic, contextual capabilities.

To address these gaps, we introduce the first \textit{conceptual taxonomy} that divides the psychotherapy process into three interconnected dimensions: Assessment, Diagnosis, and Treatment, and systematically review recent advancements and critical challenges of applying LLMs at each stage. We provide an extensive analysis of the current landscape from multiple perspectives, including the distribution of research across different psychotherapy stages, the coverage of mental disorders, the diversity of linguistic resources, and the incorporation of psychotherapy theories. Moreover, we critically evaluate the fragmented nature of existing approaches, highlighting the inadequacies in capturing dynamic symptom representations, the inherent limitations due to linguistic resource biases and problematic translations, and the diagnostic risks affecting clinical acceptance. Building on these findings, we outline essential future directions, emphasizing the need for continuous multi-stage modeling for coherent patient state tracking, real-time adaptability grounded explicitly in psychological theory, and a broadened scope of mental disorders and therapeutic frameworks. Through this comprehensive review, we aim to offer detailed methodological insights, guiding future research efforts and facilitating the practical, continuous, and theoretically-grounded integration of LLMs across the full spectrum of psychotherapy.

\paragraph{Organization of This Survey.} We present the first comprehensive survey of recent advancements in applying LLMs to psychotherapy. We introduce a conceptual taxonomy that organizes psychotherapy into three core components—Assessment, Diagnosis, and Treatment—and details their dynamic interrelations (Section \S\ref{sec:taxonomy}). We review how LLMs are applied within these components, highlighting their roles in facilitating assessments, refining diagnostic processes, and enhancing treatment strategies (Section \S\ref{sec:llms}). We examine current research trends, including symptom and language coverage as well as the distribution of various models and techniques (Section \S\ref{sec:landscape}). Finally, we discuss open challenges and outline promising directions for future work (Section \S\ref{sec:future}).





\section{Conceptual Taxonomy}
\label{sec:taxonomy}

To establish a standardized framework for understanding psychotherapy, we propose a hierarchical taxonomy aligned with the American Psychological Association (APA)'s tripartite model of psychotherapeutic processes\footnote{\url{https://www.apa.org/topics/psychotherapy}}. As illustrated in Figure \ref{fig:Taxonomy}, this taxonomy organizes psychotherapy into three core components: (1) Assessment, (2) Diagnosis, and (3) Treatment, with dynamic interconnections\footnote{Throughout this taxonomy, the terms \textit{Assessment}, \textit{Diagnosis}, and \textit{Treatment} specifically refer to the three core components of psychotherapy.}. Each component is detailed below.

\subsection{Assessment}

\paragraph{Definition.} Psychological assessment constitutes the systematic collection and interpretation of data regarding an individual's cognitive, emotional, and behavioral functioning~\cite{cohen1996psychological, kaplan2001psychological}. This process employs psychometric tests, structured clinical interviews, behavioral observations, and collateral information to establish a multidimensional profile of psychological states~\cite{groth2009handbook}.

\paragraph{Significance.} As the foundational stage of psychotherapy, assessment provides the empirical basis for understanding a client's unique psychological landscape. It enables therapists to identify symptom patterns~\cite{phillips2007assessing}, track temporal changes~\cite{barkham1993shape}, and contextualize subjective experiences within objective frameworks~\cite{groth2009handbook}. The continuous nature of psychological assessment allows for real-time adjustments to therapeutic strategies~\cite{schiepek2016real}, ensuring interventions remain responsive to evolving client needs.

\subsection{Diagnosis} 

\paragraph{Definition.} Diagnosis represents the analytical process of categorizing psychological distress using established nosological systems such as the DSM-5~\cite{Section2:11} and ICD-11~\cite{ICD11}. This involves differentiating normative emotional responses from pathological conditions while considering cultural~\cite{AlanCu2010} and developmental~\cite{kawade2012} variables that influence symptom manifestation.

\paragraph{Significance.} Diagnosis serves as the conceptual bridge between assessment and treatment, providing a structured framework for intervention planning~\cite{jensen2011understanding}. By aligning clinical observations with standardized criteria, it enhances communication among professionals~\cite{craddock2014psychiatric} and facilitates evidence-based decision-making~\cite{apa2006evidence}. 

\subsection{Treatment}

\paragraph{Definition.} Treatment includes evidence-based interventions designed to reduce psychological distress and improve functioning~\cite{apa2006evidence}. These interventions work by building a therapeutic alliance~\cite{elvins2008conceptualization}, restructuring cognition~\cite{ezawa2023cognitive}, and modifying behavior~\cite{martin2019behavior}, all typically grounded in well-established theoretical orientations.

\paragraph{Significance.} Treatment transforms the theories and information gleaned from assessment and diagnosis into practical interventions~\cite{prochaska2018systems} that directly address the client's psychological distress~\cite{barlow2021clinical} and foster personal growth~\cite{lambert2013bergin}. 

\subsection{Interrelations}
The taxonomy's components interact through three dynamic processes (see Figure \ref{fig:Taxonomy}) that define psychotherapy as a complex adaptive system:

\paragraph{Synthesizing (Assessment $\rightarrow$ Diagnosis)} The dialectical integration of observational data with nosological frameworks enables diagnostic classifications to contextualize assessment findings, \textit{synthesizing} the patient's various symptoms and behavioral patterns into a diagnostic result~\cite{rencic2016understanding}.

\paragraph{Framing (Diagnosis $\rightarrow$ Treatment)} Diagnosis functions as a \textit{framing} mechanism, integrating complex and diverse symptoms into a coherent classification that establishes a clear blueprint for treatment~\cite{Section2:11}.

\paragraph{Customization (Assessment $\rightarrow$ Treatment)}  A process where treatment plans are continuously \textit{refined} based on assessment results, considering individual differences without being constrained by diagnostic labels, to enhance therapeutic effectivenesss~\cite{Section2:5}.

\subsection{Scope of This Survey}
Recent surveys at the intersection of artificial intelligence and mental health primarily cover broad NLP-driven interventions~\cite{malgaroli2023natural} or generic AI applications in cognitive behavioral therapy~\cite{jiang2024generic}, without specific emphasis on LLMs. Other reviews explicitly focusing on LLMs, such as the scoping review~\cite{24llmmhcare,Hua2025ASR} and the overview of general opportunities and risks~\cite{lawrence2024opportunities}, examine general mental health rather than psychotherapy specifically. In contrast, our survey explicitly targets recent LLM applications within psychotherapy from the emergence of ChatGPT in late 2022 through October 2024, mainly including papers published in computational linguistics conferences and recent arXiv preprints. We adopt a slightly broad definition of LLMs, primarily including language models exceeding 7 billion parameters~\cite{peng2023does,zhao2023survey}. Using the APA’s tripartite model as a foundation, we manually classify each paper according to psychotherapy-oriented components—Assessment, Diagnosis, and Treatment--clearly highlighting critical research gaps and future directions distinct from previous reviews.

\section{LLMs in Psychotherapy}
\label{sec:llms}
\subsection{Assessment}

\begin{table*}[t!]
    \centering
    \resizebox{\textwidth}{!}{
    \begin{tabular}{c p{3.75cm} p{3.5cm} p{2.25cm} p{3.75cm}}
        \toprule
         \textbf{Study} & \textbf{Text Granularity} & \textbf{Best Technique} & \textbf{NLP Task} & \textbf{Assessment Focus} \\
         \midrule
         \multicolumn{5}{c}{\emph{Symptom Detection}}\\
         \midrule

         \citet{18} & Single Post & Emotion Prompting & BC/MCC/EG & Multiple Symptoms \\
         \citet{22} & Multi-turn Dialogue & Fine-Tuning & MLC/IE/SUM & Multiple Symptoms \\
         \citet{24} & Multi-turn Dialogue & Few-Shot Prompting & MLC/IE/SUM & PTSD \\
         \citet{27} & Single Post & Fine-Tuning & MLC/EG & Depression \\
         \citet{66} & Single Post & Few-Shot Prompting & MCC & Multiple Symptoms \\
         \citet{85} & Posts From One User & Chain-of-Thought & IE/SUM & Suicidal Ideation \\
         \citet{91} & Posts From One User & Role Prompting & IE/SUM & Suicidal Ideation \\
         \citet{94} & Single Post & Fine-Tuning & BC/MCC/EG & Multiple Symptoms \\
         \citet{95} & Single Post & Fine-Tuning & BC/EG & Multiple Symptoms \\
         \citet{96} & Single Post & Few-Shot Prompting & MLC & Multiple Symptoms \\
         \citet{108} & Single Post & Fine-Tuning & MLC & Suicidal Ideation \\
         \citet{120} & Single Post & Zero-Shot Prompting & BC & PTSD \\

         \midrule
         \multicolumn{5}{c}{\emph{Symptom Severity}}\\
         \midrule
         \citet{11} & Multi-turn Dialogue & Zero-Shot Prompting & TR & Depression/PTSD \\
         \citet{23} & Multi-turn Dialogue & Zero-Shot Prompting & TR & Depression/Anxiety \\
         \citet{58} & Posts From One User & Zero-Shot Prompting & TR & Depression \\
         \citet{59} & Posts From One User & Zero-Shot Prompting & TR & Depression \\
         \citet{100} & Single Post & Zero-Shot Prompting & TR/MCC & Depression/Suicide \\

         \midrule
         \multicolumn{5}{c}{\emph{Cognition}}\\
         \midrule
         \citet{5} & Single Sentence & Few-Shot Prompting & MLC & Cognitive Distortions \\
         \citet{10} & Single Post & Fine-Tuning & MLC & Cognitive Distortions \\
         \citet{12} & Single Sentence & Few-Shot Prompting & MCC & Cognitive Distortions \\
         \citet{14} & Single-turn Dialogue & Zero-Shot Prompting & BC/MCC/EG & Cognitive Distortions \\
         \citet{16} & Single Post & Zero-Shot Prompting & MLC & Maladaptive Schemas \\
         \citet{38} & Single Post & Zero-Shot Prompting & MCC/SUM & Cognitive Pathways \\
         \citet{45} & Single-turn Dialogue & Multi-Agent Debate & MCC & Cognitive Distortions \\

         \midrule
         \multicolumn{5}{c}{\emph{Behavior}}\\
         \midrule
         \citet{36} & Single Post & Zero-Shot Prompting & MLC/EG & Interpersonal Risk \\
         \citet{60} & Sentence From Dialogue & Few-Shot Prompting & MCC & MI-Adherent Behaviors \\
         \citet{65} & Sentence From Dialogue & Zero-Shot Prompting & MCC & MI-Adherent Behaviors \\
         \citet{67} & Sentence From Dialogue & Zero-Shot Prompting & MCC & MI-Adherent Behaviors \\

         \bottomrule
    \end{tabular}
    }
    \caption{Comparison of Psychological Assessment Studies by Input Characteristics and Methodology. \textbf{MLC}: Multi-Label Classification, \textbf{IE}: Information Extraction, \textbf{SUM}: Summarization, \textbf{MCC}: Multi-Class Classification, \textbf{BC}: Binary Classification, \textbf{TR}: Text Regression, \textbf{EG}: Explanation Generation. Studies are categorized through text granularity, optimal technical approach (\textit{Best Technique}), NLP task formulation, and specific assessment focus.}
    \label{tab:assessment_method_comparison}
\end{table*}

\paragraph{Symptom Detection} leverages LLMs to identify mental health conditions including depression, anxiety, PTSD, and suicidal ideation, demonstrating robust performance and multidimensional applicability across diverse scenarios. \citet{18} systematically evaluated GPT-3.5, InstructGPT3, and LLaMA models across 11 datasets, revealing that emotion-enhanced chain-of-thought prompting improves interpretability yet remains inferior to specialized supervised methods. \citet{22} achieved 70.8\% zero-shot symptom retrieval accuracy in Korean psychiatric interviews using GPT-4 Turbo, while their fine-tuned GPT-3.5 attained 0.817 multi-label classification accuracy. Clinical applications show particular promise, as \citet{24} leveraged GPT-4 and Llama-2 to automate PTSD assessments through information extraction from 411 interviews, significantly enhancing diagnostic practicality. 

Social media analysis benefits from approaches like \citet{27}'s interpretable depression detection framework, which demonstrated strong performance across Vicuna-13B and GPT-3.5 environments. Resource development advances include \citet{66}'s \emph{MentalHelp} dataset with 14 million instances, validated through GPT-3.5 zero-shot evaluations. For suicidal ideation monitoring, \citet{85} and \citet{91} achieved state-of-the-art evidence extraction in the CLPsych 2024 shared task through innovative prompting strategies. Open-source initiatives like \emph{MentaLLaMA} by \citet{94} and \emph{Mental-LLM} by \citet{95} enable multi-symptom detection via instruction-tuned LLaMA variants, though \citet{96}'s \emph{WellDunn} framework reveals persistent gaps in GPT-family models' explanation consistency.

Cross-lingual adaptations include \citet{108}'s \emph{PsyGUARD} system based on fine-tuned CHATGLM2-6B for Chinese suicide risk assessment, while \citet{120} demonstrated domain-specific RoBERTa models outperforming GPT-4 in cross-domain PTSD pattern analysis, highlighting the critical balance between model specialization and interpretability.

\paragraph{Symptom Severity} focuses on estimating the level of mental health condition intensity, particularly for depression, anxiety, and PTSD. Clinical evaluations reveal Med-PaLM 2's zero-shot depression scoring attains clinician-level alignment on interview data \citep{11}, though with limited PTSD generalizability. When benchmarked against specialized Transformers on DAIC-WOZ dataset~\cite{gratch-etal-2014-distress}, ChatGPT and Llama-2 exhibit moderate efficacy \citep{23}, suggesting domain-specific architectures retain advantages in structured assessments. Shifting attention to social media data, \citet{58} proposed a pipeline that retrieves depression-relevant text, summarizes it according to the Beck Depression Inventory (BDI)~\cite{jackson2016beck}, and then utilizes LLMs to predict symptom severity, achieving performance similar to expert evaluations on certain measures. In a similar vein, \citet{59} introduced an explainable depression detection system that leverages multiple open-source LLMs to generate BDI-based answers, reporting near state-of-the-art performance without additional training data. Cross-lingual extensions emerge through \citet{100}'s framework enabling severity prediction across 6 languages and 2 mental conditions.

\paragraph{Cognition} centers on identifying and understanding maladaptive thinking patterns, such as cognitive distortions and early maladaptive schemas, using LLMs. \citet{5} introduced a cognitive distortion dataset and employed a few-shot strategy with GPT-3.5 to generate, classify, and reframe them, while \citet{10} constructed two Chinese social media benchmarks for cognitive distortion detection and suicidal risk assessment, demonstrating that fine-tuned LLMs are more closely than zero-/few-shot methods to supervised baselines. In a related effort, \citet{12} released the \textit{C2D2} dataset containing 7{,}500 Chinese sentences with distorted thinking patterns.
Expanding on detection methods, \citet{14} proposed a \textit{Diagnosis of Thought (DoT)} prompting approach for GPT-4 and ChatGPT, which breaks down patient utterances into factual versus subjective content and supports the generation of interpretable diagnostic reasoning. Beyond cognitive distortions, \citet{16} investigated zero-shot approaches with GPT-3.5 to identify early maladaptive schemas in mental health forums, highlighting challenges in label interpretability and prompt sensitivity. Complementarily, \citet{38} presented a hierarchical classification and summarization pipeline to extract cognitive pathways from Chinese social media text, underscoring GPT-4's strong performance albeit with occasional hallucinations. Finally, \citet{45} introduced a multi-agent debate framework for cognitive distortion classification, reporting substantial gains in both accuracy and specificity by synthesizing multiple LLM opinions before forming a final verdict.

\paragraph{Behavior} highlights how user actions--or in the case of Motivational Interviewing (MI), language itself--can serve as a measurable indicator of one's readiness for change. For instance, \citet{36} introduced the \textit{MAIMS} framework, employing mental scales in a zero-shot setting to identify interpersonal risk factors on social media, thereby enhancing both interpretability and accuracy. In clinical dialogues, \citet{60} demonstrated how LLMs can automatically detect a client's motivational direction (e.g., change versus sustain talk) and commitment level, offering valuable insights for MI-based interventions. Extending such analyses to bilingual settings, \citet{65} proposed the \textit{BiMISC} dataset and prompt strategies that enable LLMs to code MI behaviors across multiple languages with expert-level performance. Lastly, \citet{67} presented \textit{MI-TAGS} for automated annotation of global MI scores, illustrating how context-sensitive modeling can approximate human annotations in psychotherapy transcripts.

\paragraph{Advanced} research has evolved beyond foundational assessment tasks to emphasize novel methodological paradigms, bias mitigation, and domain-specific summarization frameworks. For instance, \citet{48} introduced \emph{PsychoGAT}—an interactive, game-based approach that transforms standardized psychometric instruments into engaging narrative experiences, improving psychometric reliability, construct validity, and user satisfaction when measuring constructs such as depression, cognitive distortions, and personality traits. In parallel, \citet{53} systematically investigated potential biases in various LLMs across multiple mental health datasets, revealing that even high-performing models exhibit unfairness related to demographic factors. The authors proposed fairness-aware prompts to substantially reduce such biases without sacrificing predictive accuracy. Furthermore, \citet{99} presented the \emph{PIECE} framework, which adopts a planning-based approach to domain-aligned counseling summarization, structuring and filtering conversation content before integrating domain knowledge.


\subsection{Diagnosis}

\paragraph{Static Diagnosis} is based on a fixed set of data, typically derived from complete dialogues or social media posts. \newcite{11} highlighted the effectiveness of Med-PaLM 2 in psychiatric condition assessment from patient interviews and clinical descriptions without specialized training. Similarly, \newcite{107} showcased LLMs' superior performance on depression and anxiety detection on Russian datasets, particularly with noisy or small datasets. \newcite{102} evaluated PLMs and LLMs on multi-label classification in depression and anxiety, underscoring the ongoing challenges in applying LMs to mental health diagnostics. Besides, \newcite{124} introduced \textit{DORIS}, a depression detection system integrating text embeddings with LLMs, utilizing symptom features, post-history, and mood course representations to make diagnostic predictions and generate explanatory outputs. \newcite{110} developed \textit{ADOS-Copilot} for ASD diagnosis through diagnostic dialogues, employing In-context Enhancement, Interpretability Augmentation, and Adaptive Fusion based on real-world ADOS-2 clinic scenarios.

\paragraph{Dynamic Diagnosis} involves real-time evaluation based on ongoing, interactive conversations between the patient and LLM, enabling more personalized and contextually relevant insights. \newcite{9} simulated psychiatrist-patient interactions with ChatGPT, in which the doctor chatbot focused on role, tasks, empathy, and questioning strategies, while the patient chatbot emphasized symptoms, language style, emotions, and resistance behaviors. \newcite{33} introduced the \textit{Symptom-related and Empathy-related Ontology (SEO)}, grounded in DSM-5 and Helping Skills Theory, for depression diagnosis dialogues. \newcite{52} dissected the doctor-patient relationship into psychologist’s empathy and proactive guidance and introduced \textit{WundtGPT} that integrated these elements. \newcite{106} further presented the \textit{AMC}, a self-improving conversational agent system for depression diagnosis through simulated dialogues between patient and psychiatrist agents.

\subsection{Treatment}

\paragraph{LLM as a Virtual Therapist} centers on leveraging LLMs to directly engage in therapeutic conversations, often adopting multi-turn dialogues that incorporate recognized psychotherapeutic frameworks. For instance, \citet{20} proposed \emph{HealMe} to facilitate cognitive reframing and empathetic support in line with established psychotherapy principles. Likewise, \citet{29} introduced \emph{CaiTI}, a system embedded in everyday smart devices that conducts assessments of users’ daily functioning and delivers psychotherapeutic interventions through adaptive dialogue flows. In a similar vein, \citet{40} presented \emph{CoCoA}, specializing in identifying and resolving cognitive distortions via dynamic memory mechanisms and CBT-based strategies, while \citet{54} proposed a step-by-step approach guiding users to execute self-guided cognitive restructuring through multiple interactive sessions. Beyond standard CBT protocols, \citet{55} focused on aiding psychiatric patients in journaling their experiences, thereby offering richer clinical insights, whereas \citet{77} developed a multi-round CBT dataset to refine LLMs for direct counseling-like interactions. Additionally, multi-agent frameworks like \emph{MentalAgora} \citep{78} highlighted personalized mental health support by integrating multiple specialized agents, and \citet{109} further explored “mixed chain-of-psychotherapies” to combine various therapeutic methods, aiming to enhance the emotional support and customization delivered by chatbot interactions.

\paragraph{LLM as an Assistive Tool} refrains from providing a holistic therapy role but instead offers targeted support such as rewriting suboptimal counselor responses, generating controlled reappraisal prompts, or aiding clinicians in specific tasks. For example, \citet{2} proposed to rewrite responses that violate MI principles into MI-adherent forms, ensuring more consistent therapeutic dialogue. Meanwhile, \citet{3} and \citet{5} focused on generating single-turn reframes of negative thoughts--often anchored in cognitive distortions--through controlled language attributes. On the detection side, \citet{35} built a multimodal pipeline to identify depression and provide CBT-style replies, albeit with an emphasis on technological assistance rather than full-fledged therapy. In the Chinese context, \citet{43} combined cognitive distortion detection with ``positive reconstruction,'' demonstrating a single-round rewrite approach for negative or distorted statements, while \citet{64} showcased a structured Q\&A format that offers professional yet succinct CBT-based responses. From a knowledge-distillation angle, \citet{70} demonstrated how smaller models could replicate GPT-4's MI-style reflective statements, and \citet{81} introduced a lighter-weight framework \emph{RESORT} to guide smaller LLMs toward effective cognitive reappraisal prompts, thus enabling broader accessibility of self-help tools.

\paragraph{LLM as Simulated Patients for Clinician Education} pivots toward generating synthetic yet realistic patient behaviors or multi-level feedback to train or support mental health practitioners. For instance, \citet{49} leveraged LLMs to deliver multi-tier feedback on novice peer counselors’ conversational skills, significantly reducing the need for continuous expert oversight. Similarly, \citet{51} introduced LLM-driven patient simulations that help trainees practice CBT core skills in a controlled, repeatable setup. In the realm of assessing therapy quality, \citet{56} showcased a digital patient system to evaluate MI sessions, employing AI-generated transcripts to differentiate novice, intermediate, and expert therapeutic skill levels. Complementarily, \citet{80} offered \emph{Roleplay-doh}, a pipeline wherein domain experts craft specialized principles that guide LLM-based role-playing agents, thereby providing customizable training for new therapists.

\begin{figure*}[t]
\centering
\includegraphics[width=1\linewidth]{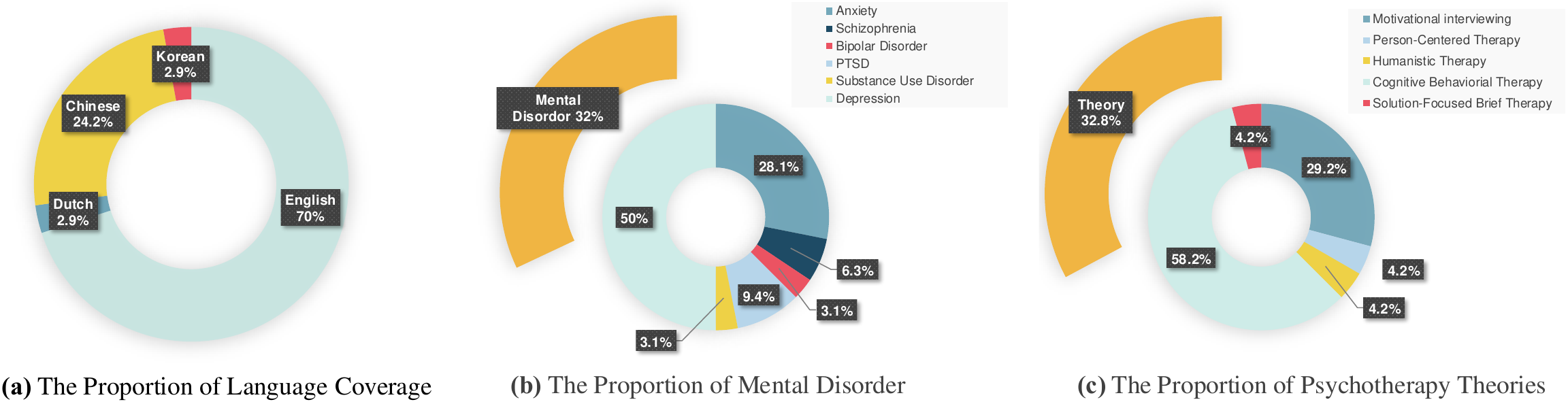}
    \caption{Distribution Analysis of The Current Landscape. Panel (a) indicates English is the predominant language (70\%), with Chinese also represented (24.2\%). Panel (b) shows that 32\% of studies address mental disorders, with Depression and Anxiety being the most common topics within this group. Panel (c) reveals that 32.8\% of studies incorporate psychotherapy theories, where CBT is the most frequently applied.}
    \label{fig:data}
\end{figure*}

\paragraph{LLM for Evaluation and Quality Analysis} targets the appraisal of therapy dialogue, counselor techniques, and treatment processes, typically without delivering direct interventions to clients. For instance, \citet{4} augmented crisis counseling outcome prediction by fusing annotated counseling strategies with LLM-derived features, achieving substantially improved accuracy. In the Chinese context, \citet{19} introduced \emph{CPsyCoun}, employing reports-based dialogue reconstruction and automated evaluation to verify counseling realism and professionalism. Beyond single-session analyses, \citet{32} used simulated clients to assess perceived therapy outcomes, while \citet{37} created the \emph{BOLT} framework for systematically comparing LLM-based therapy behaviors with high- and low-quality human sessions. Further extending to online counseling, \citet{39} proposed an LLM-based approach to measure therapeutic alliance, whereas \citet{62} delineated therapist self-disclosure classification as a new NLP task. In the MI domain, \citet{65} and \citet{67} collected bilingual transcripts to systematically annotate therapist–client exchanges for behavior coding and global scores, respectively. Additionally, multi-session perspectives emerge in \citet{101}, who proposed \emph{IPAEval} to track long-term progress from the client’s viewpoint, and \citet{103} analyzed conversation redirection and its impact on patient–therapist alliance over multiple sessions. Finally, \citet{111} and \citet{122} explored the disparities between LLM- and human-led CBT sessions, highlighting gaps such as empathy and cultural nuance while also introducing \emph{CBT-Bench} to probe LLMs’ deeper psychotherapeutic competencies.

\section{Current Landscape}
\label{sec:landscape}

\subsection{Overview}
Our survey encompasses a total of 69 studies in the field of LLMs in psychotherapy. Specifically, 33 studies address assessment, 9 focus on diagnosis, and 32 concentrate on treatment, with 5 studies overlapping across these dimensions. Approximately 74\% of the studies employed commercial large language models, while about 77\% used prompt-based techniques. This distribution highlights an imbalance in research focus across different stages of the psychotherapy process and reflects a heavy reliance on commercial models and prompt technologies.

Figure \ref{fig:data} presents a comprehensive analysis of the current research landscape in this field. 
Panel \textbf{(a)} reveals a significant linguistic bias in existing studies, with English-language corpora dominates. 
While there are limited studies involving Korean and Dutch languages, this highlights a substantial gap in multilingual research approaches.
Panel \textbf{(b)} quantitatively demonstrates the distribution of mental health research focuses. Mental disorder-related studies constitute 32\% of the total research corpus (represented by the orange outer ring).
Within this subset, depression-focused research accounts for 50\% of mental disorder studies, followed by anxiety-related research. This distribution indicates a concerning imbalance, where common conditions receive disproportionate attention while more complex disorders, such as bipolar disorder, remain understudied.
The analysis of psychotherapy theories in panel \textbf{(c)} uncovers another critical gap in the field. Only 32.8\% of the studies incorporate psychotherapy theories in their methodological approach. 
Notably, emerging therapeutic frameworks, such as humanistic therapy, are particularly underrepresented in current research applications.

\subsection{Fragmented Approaches}
Traditionally, LLM-based psychotherapy tools have addressed assessment, diagnosis, and treatment separately. Recently, a few studies have started to explore integrative approaches spanning multiple stages. Despite these emerging integrative efforts, the systems remain limited, typically addressing only two stages without achieving full continuity. Additionally, fragmentation occurs not only across these three dimensions but also at more granular levels; for example, some methodologies are narrowly focused on assessing single disorders~\cite{24,27}, further limiting their applicability and integration potential in broader psychotherapy contexts.

\subsection{Critical Issues and Risks}
\paragraph{Dynamic Symptom Representation.} Psychotherapy commonly involves shifting symptoms, comorbidities, and nuanced patient experiences, making static or single-label predictions insufficient. Current models fail to adequately capture multi-label conditions and temporal symptom fluctuations, leading to incomplete or inaccurate assessments~\cite{lee-etal-2024-detecting-bipolar}.

\paragraph{Linguistic Resource Bias and Translation Critique.} Most psychotherapy-oriented LLMs are trained primarily on English datasets, with some researchers attempting to expand linguistic coverage through translation. However, recent studies highlight significant cultural specificity in mental health disorders~\cite{watters2010crazy, abdelkadir-etal-2024-diverse}, making direct translation of datasets unreliable for accurately capturing psychological nuances across different cultures.

\paragraph{Diagnostic Risks.} Current approaches to automated diagnosis often struggle to gain acceptance among clinical practitioners due to concerns about reliability and patient safety. Despite the psychology community increasingly favoring transdiagnostic methods~\cite{dalgleish2020transdiagnostic}, a segment of NLP researchers continues to emphasize diagnosis-specific studies, indicating a notable divergence in research priorities.

\section{Future Directions}
\label{sec:future}
\paragraph{Continuous Multi-Stage Modeling.} Psychotherapy inherently involves continuous interactions that progress through assessment, diagnosis, and intervention phases. Existing research indicates that several leading foundation models exhibit negligible hallucination issues in the medical domain~\cite{kim2025medical}, providing a promising foundation for integrating these stages, as minimizing hallucinations is crucial for ensuring the accuracy and reliability required for continuous patient state tracking across psychotherapy stages. Future models should aim for an evolving representation of patient states, ensuring consistency and coherence across the entire therapeutic process rather than handling stages as isolated segments.

\paragraph{Real-Time Adaptability Grounded in Psychological Theory.} The development of real-time adaptive strategies represents a significant step beyond current static models. Current technical advancements, such as retrieval-augmented generation and long-context memory techniques~\cite{jo2024understanding}, provide the necessary technical foundations for such strategies. Instead of simply reproducing patterns found in pre-collected dialogue datasets, future LLM applications should incorporate these advanced contextual memory mechanisms informed by established psychological theories. Such systems would dynamically interpret patient cues, adjusting interventions immediately in response to subtle shifts in emotional and cognitive states. This approach significantly surpasses mere language-style mimicry achieved through simple fine-tuning on existing datasets, enabling deeper, theoretically-informed therapeutic engagement.

\paragraph{Broadening Scope of Disorders and Therapeutic Approaches.} Future research should prioritize diversification in terms of both disorders addressed and therapeutic methodologies employed. The current focus on common disorders like anxiety and depression has led to an imbalanced research landscape. There is a pressing need to incorporate underrepresented conditions, such as bipolar disorder and personality disorders, alongside a broader spectrum of psychotherapeutic frameworks, including psychodynamic~\cite{shedler2010efficacy} and existential-humanistic approaches~\cite{schneider2010existential}. Such expansion would help address existing blind spots, contributing to a more inclusive and comprehensive application of LLMs in psychotherapy.

\section{Conclusion}
LLMs hold significant promise for revolutionizing psychotherapy by enhancing assessment, diagnosis, and treatment. However, the current landscape reveals critical limitations: research is often fragmented across these stages, exhibits notable biases in linguistic and disorder coverage, and underutilizes diverse psychotherapeutic theories. To overcome these challenges, future work must focus on creating continuous, multi-stage models that are grounded in psychological theory and capable of real-time adaptation to evolving patient states. Expanding the scope of addressed disorders and therapeutic modalities will be crucial for developing LLM-driven psychotherapy tools.

\section*{Limitations}
We remind the readers that this survey has the following limitations: 
1) The studies reviewed primarily focus on the application of LLMs in psychotherapy, and there may be relevant research in adjacent fields or interdisciplinary domains that was not included. 
2) Due to the rapidly evolving nature of this area, some recent advancements may not be captured. The scope of this survey is limited to the available literature and may overlook emerging trends or unpublished findings. 
3) While we provide a taxonomy of LLM applications in psychotherapy, this framework may not fully encompass the complexity of real-world clinical settings or the diverse range of therapeutic approaches currently in practice.

\bibliography{custom}

\appendix



\end{document}